\documentclass{article}

 \usepackage[nonatbib, final]{nips_2017}

\usepackage[utf8]{inputenc} 
\usepackage[T1]{fontenc}    
\usepackage{hyperref}       
\usepackage{url}            
\usepackage{booktabs}       
\usepackage{amsfonts}       
\usepackage{nicefrac}       
\usepackage{microtype}      

\usepackage{graphicx}
\usepackage{graphics}
\usepackage{float}
\usepackage{caption}

\usepackage{amsthm}
\usepackage{amsmath}
\usepackage{amssymb}
\usepackage{dsfont}
\usepackage{mathrsfs}
\usepackage[amssymb]{SIunits}
\usepackage[mathscr]{eucal}

\usepackage[numbers]{natbib}

\title{Generative Adversarial Networks for Electronic Health Records: A Framework for Exploring and Evaluating Methods for Predicting Drug-Induced Laboratory Test Trajectories}

\author{
	  Alexandre Yahi \quad Rami Vanguri \quad No\'emie Elhadad* \quad Nicholas P. Tatonetti*\\
	  Department of Biomedical Informatics, Columbia University, New York, USA.\\
	\texttt{\{alexandre.yahi, r.vanguri, noemie.elhadad, nick.tatonetti\}@columbia.edu} \\
	* co-senior authors
}
\DeclareMathOperator*{\argmin}{arg\,min}

\begin{document}

\maketitle

\begin{abstract}
Generative Adversarial Networks (GANs) represent a promising class of generative networks that combine neural networks with game theory. From generating realistic images and videos to assisting musical creation, GANs are transforming many fields of arts and sciences. However, their application to healthcare has not been fully realized, more specifically in generating electronic health records (EHR) data. In this paper, we propose a framework for exploring the value of GANs in the context of continuous laboratory time series data. We devise an unsupervised evaluation method that measures the predictive power of synthetic laboratory test time series. Further, we show that when it comes to predicting the impact of drug exposure on laboratory test data, incorporating representation learning of the training cohorts prior to training GAN models is beneficial.    
\end{abstract}

\section{Introduction}
Since the development of the Generative Adversarial Networks (GANs) framework described by \citet{goodfellow2014generative} in 2014, about 188 publications as of September 2017 have presented  novel variations, architectures, optimization algorithms and applications for this class of methods. GANs are generative models where two neural networks, a Discriminator $D$ and a Generator $G$ compete in a zero-sum game theory context. The goal of $D$ is to correctly discriminate real samples from synthetic samples created by $G$. The goal of $G$ is to create realistic samples that decrease D's accuracy until they reach a Nash equilibrium where the Discriminator cannot do better than guessing if samples are real or not. While GANs have demonstrated their power in applications such as text-to-image synthesis \cite{reed2016generative, zhang2016stackgan}, image-to-image translation \cite{isola2016image}, video generation \cite{vondrick2016generating}, and even music generation \cite{yang2017midinet}, applications to healthcare remain scarce.

Besides use cases in medical imaging such as SegAN presented by \citet{xue2017segan} for medical image segmentation, two GANs architectures applied to electronic health records (EHR) stand out: the RCGAN, a Recurrent Conditional GAN capable of generating real-values time series evaluated with supervised learning tasks \cite{esteban2017real} and medical GAN (medGAN)\cite{choi2017generating}, an algorithm that can generate synthetic Electronic Health Records (EHR) matrices of binary or count features using an autoencoder to learn latent features and force the outputs to discrete variables.

Deep generative models represent an opportunity in biomedical sciences for various applications. They have been identified as a promising method to release de-identified biomedical data that can be shared while preserving privacy \cite{beaulieu2017privacy, choi2017generating} and can support model learning in supervised setups \cite{esteban2017real}. GANs also have shown potential for prediction and inference outside of the biomedical realm, with papers using the GAN frameworks to predict video frames \cite{mathieu2015deep} and others performing adversarially learned inference \cite{dumoulin2016adversarially}. There is little precedent for trying to forecast laboratory test trajectories in an unsupervised fashion. The benefits of predicting patient-specific laboratory test trajectories could be significant in drug safety in particular, where laboratory tests and other temporal measurements represent reliable biomarkers for drug-drug interactions adverse drug reactions, as shown by \citet{tatonetti2011detecting} with Paroxetine and Pravastatin increasing blood glucose levels and \citet{lorberbaum2016coupling} with drug-induced QT prolongation. Being able to train drug-specific GAN models would inform the detection of such events or even quantify the impact of a drug on a particular laboratory test.

Our objective in this paper is to generate continuous time series that display effects of exposure changes with a simple GAN architecture. More specifically, we focused on drug laboratory effects (DLEs) -- decrease or increase of laboratory test due to a specific drug exposure -- in order to compare real and synthetic time series and their effects on the statistics of pre and during exposure measurements. To this end, we decided to study the effect of HMG-CoA reductase inhibitors, or statins, on cholesterol laboratory measurements they are designed to decrease \cite{istvan2001structural}. Our cohort of interest is therefore exclusively composed of patients exposed to statins. We decided to leave the inclusion of cholesterol measurements from patients not exposed to statins for future work and focus here only on the exposure effects. We are aware of the extensive use of recurrent neural networks on time series in the literature, but having aligned all the time series in our experiments did not justify their use for the present study.

EHR data are known to be complex due to their multi-modality \cite{hripcsak2012next}, mixing categorical and continuous data with semi-structured and free text medical notes. There is no study to our knowledge that investigates the impact of that complexity when training generative models and how clinically meaningful cohort stratification can improve the accuracy of synthetic data. The key contributions of this paper are: (1) Describing a method to normalize laboratory test time series to study drug laboratory effects (DLEs); (2) Demonstrating that clinically driven deep cohort stratification results in more accurate GANs; (3) Proposing an unsupervised evaluation method to GAN models by measuring the predictive power of synthetic laboratory test series.

\section{Methods}
\subsection{Data}
Our electronic health records (EHR) data were collected at the New York Presbyterian/Columbia University Irving Medical Center between 2000 and 2013 with 19.6 million drug prescriptions for 485,306 patients and 473.6 million laboratory test observations. We selected all the patients exposed to HMG-CoA reductase inhibitors, or statins (ATC code C10AA) at any point in time. For each of these patients we collected all the total cholesterol measurements (LOINC code 2093-3) they had.

\subsection{Pre-processing}
For each patient exposed to statins, we annotated the total cholesterol measurements to determine if they were falling inside or outside a window of continous exposure to a statin with no gap larger than 30 days. We split these timelines on and off drug into segments of measurements during an exposure era preceded by measurements off-exposure at most a year before the beginning of the drug era. Exposure eras without pre-exposure data following these criteria were excluded. In order to feed our neural networks vectors of fixed length, we performed a linear interpolation weighted by the measurement dates to have the same number of points before and during drug exposure. In order not to distort the pattern between before and during drug exposure, we made sure to interpolate independently the pre-exposure measurements and the during exposure measurements. Finally, for better training performances with our GANs architecture, we linearly normalized between -1 and 1 all these time series using the 99th percentile values of total cholesterol measurements at any point in time for patients exposed to statins, which removed potentially erroneous values and extreme outliers. Values exceeding this range were brought back to -1 if too low and 1 if too high.

\subsection{Deep patients stratification}
We collected clinical covariates for our cohort in the period before drug exposure: drug prescriptions in the form of ATC codes and ICD-9 diagnoses code that we grouped by 3-digit codes. We trained a deep autoencoder with four encoding layers (256, 128, 64, 32) and four decoding layers on these binary features before exposure. This approach is analogous to the \textit{Deep Patient} model \cite{miotto2016deep}. We then applied the t-SNE algorithm developed by \citet{maaten2008visualizing} to represent patients in two dimensions and clustered with spectral clustering.

\subsection{Generative Adversarial Network training}
We used the mini-batch averaging method proposed by \citet{choi2017generating} with some notable differences: the model was performing best when the autoencoder's encoder layer had the same size as the input, therefore not compressing but simply identifying meaningful features. We selected a tanh reconstruction function and evaluated the loss with a mean square error given the continuous nature of our normalized time series. The generator had a single layer of size 16, and the discriminator had two layers of sizes 32 and 16. Given the low number of features that we had, we avoided over-fitting the model by having too many hidden units. Each GAN was trained on 100 epochs and with a batch size of 10, except for the smallest cluster that was trained with a batch size of 5.

\subsection{Evaluation of GAN outputs}

Our evaluation process was 2-tiered: first, we evaluated the presence and magnitude of the drug effect using a paired t-test on the average values before and during drug exposure for each measurement segment. We defined the effect size as the mean of the differences between both groups of measurements.
Second, we evaluated how predictive synthetic time series were as a measure of fidelity to the real data. To do so, we used the mean square error $\text{MSE}_{pre}(x,\hat{x})$ as a measure of similarity between a real time series $x$ and a synthetic one $\hat{x}$ using the before exposure values only. The predictivity error was expressed as the mean square error $\text{MSE}_{exp}(x,\hat{x})$ on the values during exposure only. To evaluate the predictivity of a set $\hat{S}_V$ of $V$ synthetic series for a given set $S_N$ of $N$ real series, we expressed the predictivity error $P_{err}$ as:
\[ P_{err}(S_N,\hat{S}_V) = \dfrac{1}{N} \sum_{k=1}^{N} \text{MSE}_{exp}(x_k,\argmin_{\hat{x}_j \in V}(\text{MSE}_{pre}(x_k,\hat{x}_j))  \]

which is the mean of the predictivity errors from each real series $x_k$ with its most similar synthetic series based on the pre-exposure measurements.
For each cluster identified, we compared  $P_{err}$ between the GAN trained on the whole cohort, and the GAN trained on the cluster of interest. For each GAN model we generate 10 times more synthetic samples than the largest cluster, and then sampled as many synthetic time series as there are real ones. The random clusters were obtained with sampling without replacement and were the same size as the real clusters for each test.

\section{Results}

We identified 65,563 patients exposed to any statin in the inpatient setting, with a total of 411,880 total cholesterol measurements (average per patient: 8.4, min:1, max:313). After pre-processing their total cholesterol measurement series, and excluding patients with no measurement during drug exposure and patient without measurements within a year before exposure, we ended up with 4,830 patients (50.6\% females) with an average age at statin exposure of 65.15 (SD: 12.25). Each patient was associated with one interpolated time series of total cholesterol of 16 points (8 measurements before, 8 measurements during exposure). No patient had enough data to have more than one before/during exposure measurement series according to our criteria. Our cohort presented an average total cholesterol value of 170.6 mg/dL (SD: 56.79) before statin exposure, and 160.7 mg/dL (SD: 50.98) during statin exposure, significantly lower ($p<10^{-79}$), compared to the average of 185.3 mg/dL (SD: 50.48) for the 1,345,017 values available in our EHR system. The 99 percentile interval of total cholesterol for this cohort was $[75-319]$ mg/dL.

By compressing the 1524 clinical features into 32 dimensions with our deep autoencoder and representing them in two dimensions using t-SNE, we identified two clearly defined clusters and one larger cluster that the spectral clustering cut in two. Remarkably, cluster 3 was solely composed of type 2 diabetes patients. We trained GANs on the total cohort (totalGAN) and on each cluster (subGANs) and represented the average synthetic time series they generated compared to the real one with their standard deviation (Figure 1). We then evaluated the accuracy of subGANs against the totalGAN using the similarity and prediction MSE described in the Methods section above.
\begin{figure}[h]
  \centering
	\includegraphics[scale=0.65]{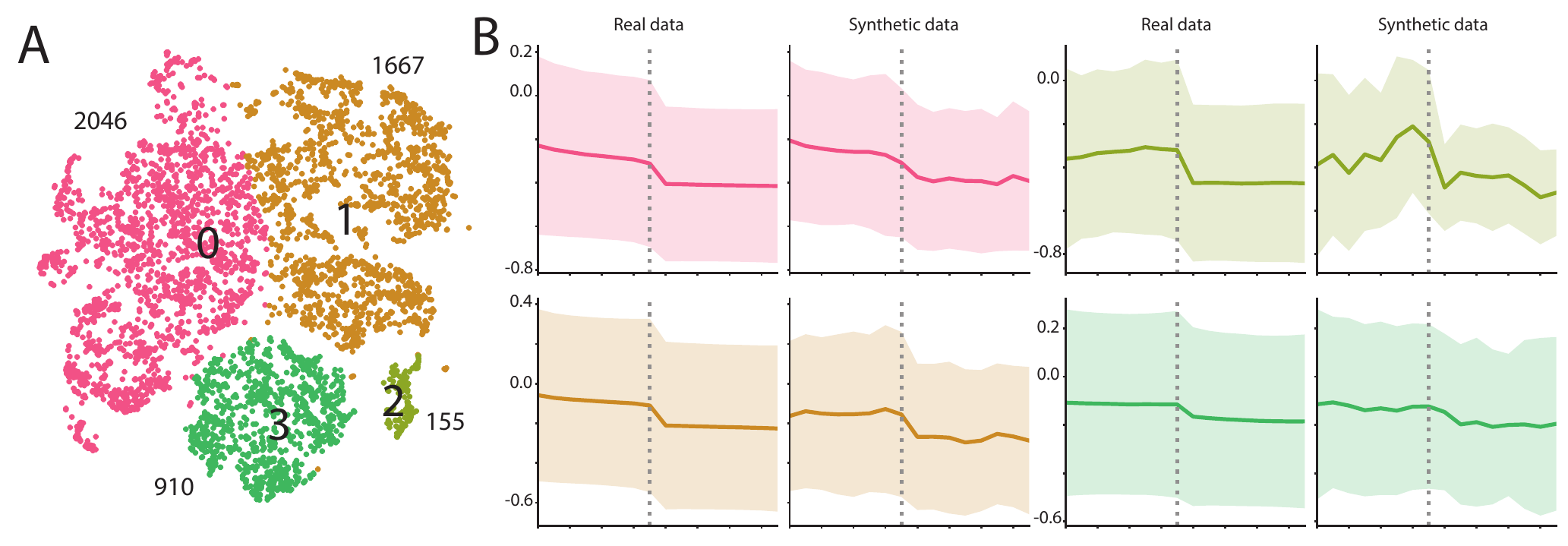}
  \caption{A. Clustered t-SNE representation of the time series encoded using clinical co-variates B. Comparison of the average time series generated by the GANs trained on each cluster with the real time series.}
  \vspace*{-1em}
\end{figure}

\begin{table}[h]
  \caption{Predictivity error $P_{err}$ ($\pm$ SD)}
  \label{sample-table}
  \centering
  \begin{tabular}{lllllll}
 	\midrule
    & \multicolumn{3}{c}{Clinical Clusters}
    &
    \multicolumn{3}{c}{Random Clusters} \\
    \cmidrule(r){2-4}\cmidrule(l){5-7}
    Cluster     & subGAN  & totalGAN & p-value & subGAN & totalGAN & p-value\\
    \midrule
    Cluster 0 & 0.13 ($\pm$0.22)  &  0.16 ($\pm$0.28)  & 5.7e-4 & 0.27 ($\pm$0.39) & 0.16 ($\pm$0.27) & 2.0e-303 \\
   	Cluster 1 & 0.15 ($\pm$0.27) & 0.16 ($\pm$0.26) & 1.5e-1 & 0.30 ($\pm$0.45) & 0.16 ($\pm$0.26) & 2.2e-308  \\
   	Cluster 2 & 0.11 ($\pm$0.21) & 0.22 ($\pm$0.35) & 3.9e-5 & 0.24 ($\pm$0.40) & 0.15 ($\pm$0.26) & 1.1e-21  \\
    Cluster 3 & 0.12 ($\pm$0.20) & 0.15 ($\pm$0.24) & 3.9e-4 & 0.28 ($\pm$0.38) & 0.16 ($\pm$0.27) & 1.5e-144  \\
    \bottomrule
  \end{tabular}
\end{table}

The real data of each cluster was significantly better predicted by the subGANs than by the totalGAN for all clusters except for Cluster 1. Moreover, subGANs trained on random clusters of identical size significantly performed worse than the clinically relevant ones (Table 1), hinting at the importance of clinical variables.

\section{Discussion}

In this paper we presented an unsupervised framework to evaluate generative adversarial networks for the prediction of drug-induced laboratory test trajectories. This framework is applicable to any time series affected by a known exposure factor. We defined a similarity measure to align synthetic time series to real ones before exposure and a metric to evaluate the prediction performances of synthetic time series during the exposure period. Further, we demonstrated that clinical variables can be integrated to identify meaningful clusters that produce significantly more accurate GANs with regard to exposure trajectory prediction. By using a deep autoencoder, we hint at the potential for integrating neural network-based compressed representations of patients into conditional GANs architectures while keeping evaluations completely unsupervised. Such architectures would also direct the synthetic data generation to ensure an increased similarity with the real data. Finally, more work needs to be conducted to evaluate how to sample efficiently from the learned distributions to go further in the trajectory prediction process and provide probability densities rather than a prediction error.

\section{Conclusion}

In conclusion, we presented a novel unsupervised framework for evaluating the use of generative adversarial model on clinical time series and the prediction of their trajectory after a known exposure. In future work, we will integrate the patient deep representation into the GAN architecture to improve predictive power.

\bibliographystyle{plainnat}
\bibliography{nips17_ay_bib} 

\appendix

\section{Total cholesterol distributions}
In this section, we provide the total cholesterol distributions' statistics and figures for the different cohorts mentions in the Results section. Normal ranges for total cholesterol (LOINC 2093-3) are below 200 mg/dL. Between 200 and 240 mg/dL the level is considered to be borderline high. Total cholesterol is considered to be high for values above 240 mg/dL. As mentioned in the Results section, we collected a total of 411,880 total cholesterol measurements in the period of interest of our retrospective study. The minimal value measured was 0.0 and the maximal value was 3368 mg/dL, yielding an average of 184.6 mg/dL ($\pm$ 45.58). To remove outliers in that study, we only considered the 99 percentiles as represented in Figure S1 showing the density estimate distribution of total cholesterol values over all patients between 2000 and 2013.

\begin{figure}[h]
	\centering
	\includegraphics[scale=0.65]{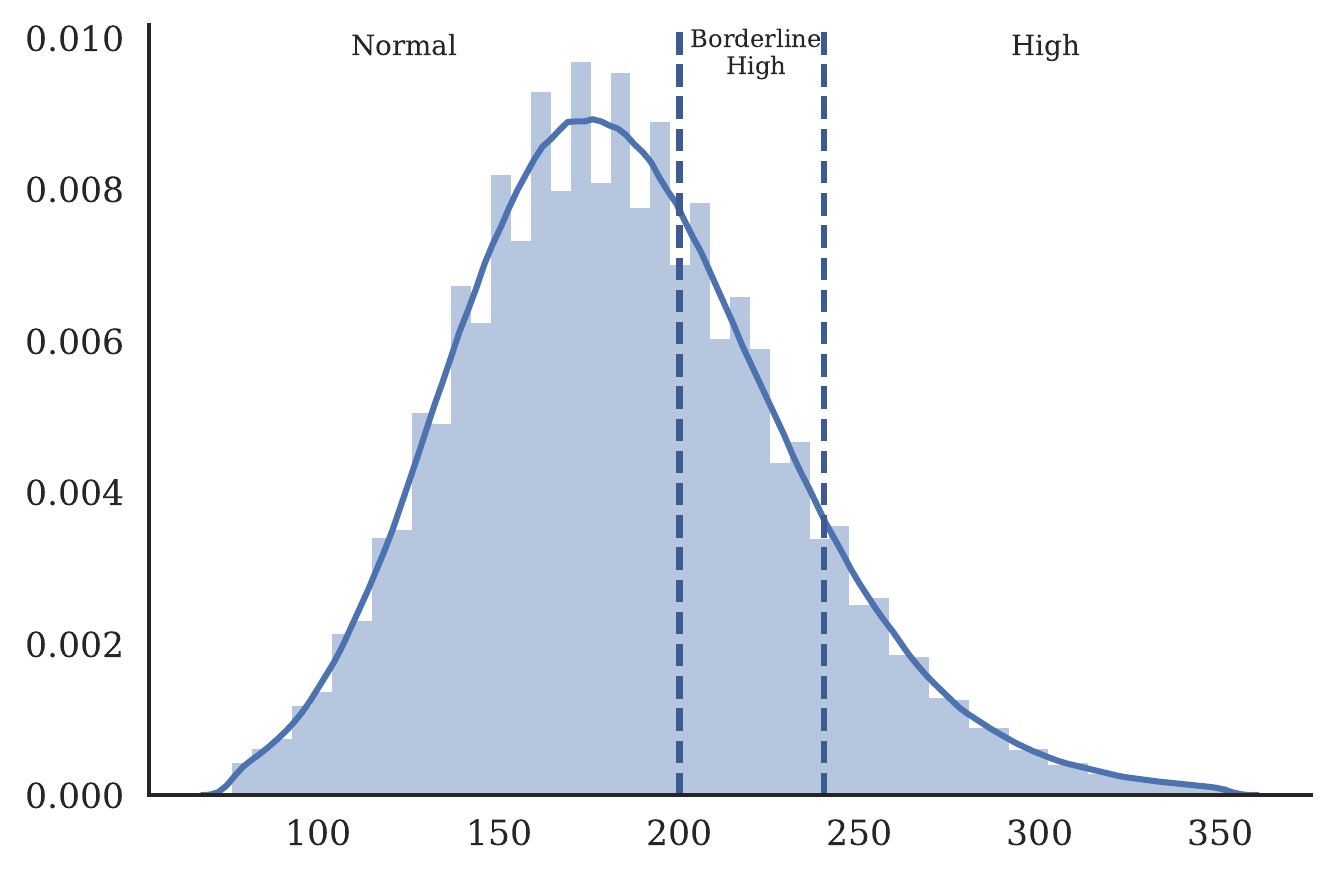}
	\caption*{Figure S1. 99 percentiles of the total cholesterol distribution for all patients in our clinical data warehouse between 2000 and 2013, with total cholesterol (LOINC 2093-3) normal ranges.}
	\vspace*{-1em}
\end{figure}

Figure S2. represents the laboratory test measurements after linear interpolation and linear normalization using the 99 percentile extremal values as normalization bounds, for the complete cohort for all time points, for the measurements before exposure, and the measurement during exposures, and the four sub clusters identified using clinical variables as described in the Methods section 2.3.

\begin{figure}[h]
	\centering
	\includegraphics[scale=0.69]{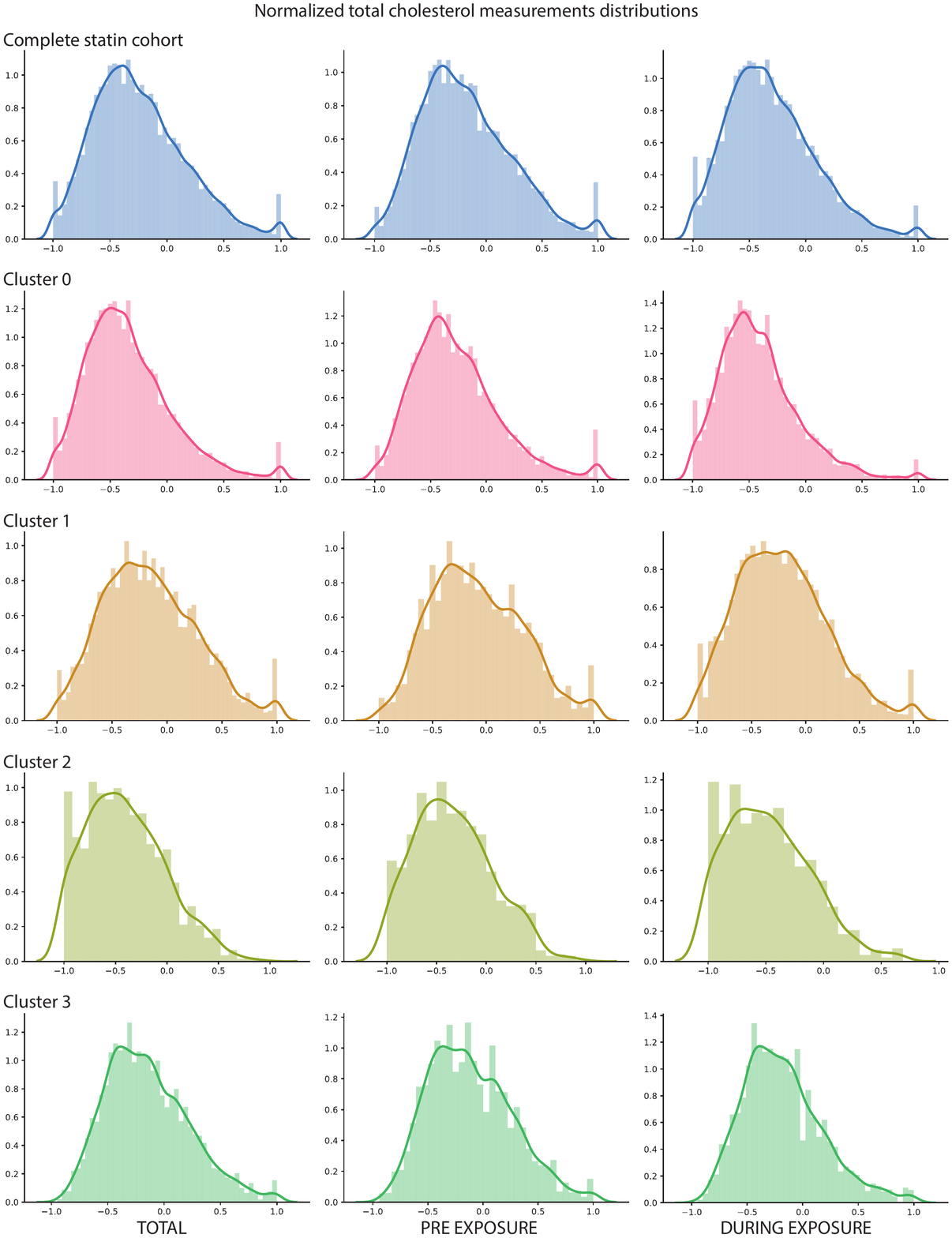}
	\caption*{Figure S2. Normalized total cholesterol measurements for the complete statin cohort, and the 4 sub-clusters, overall (TOTAL), before exposure, and during exposure to statins.}
	\vspace*{-1em}
\end{figure}

\section{Generated time series figures}

In this subsection, we present randomly selected real time series from each of the four sub clusters with the synthetic time series from the GAN trained on the whole cohort (totalGAN), and the GAN trained on the specific cluster (subGAN)  that were the closest before exposure to provide visualize cues about the predictive errors (Figure S3-6).

\begin{figure}[h]
	\centering
	\includegraphics[scale=0.50]{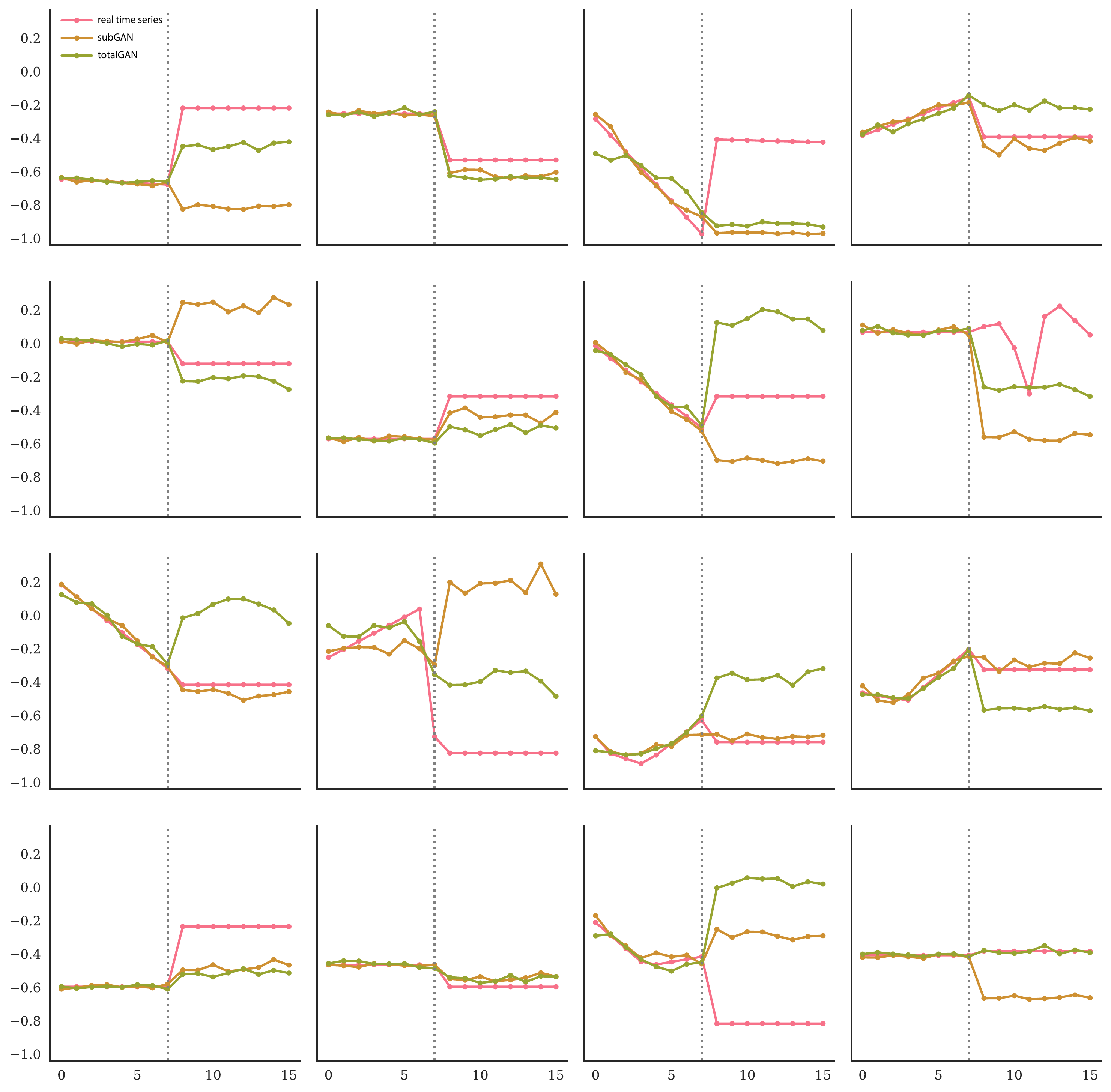}
	\caption*{Figure S3. Cluster 0.}
\end{figure}

\begin{figure}[h]
	\centering
	\includegraphics[scale=0.50]{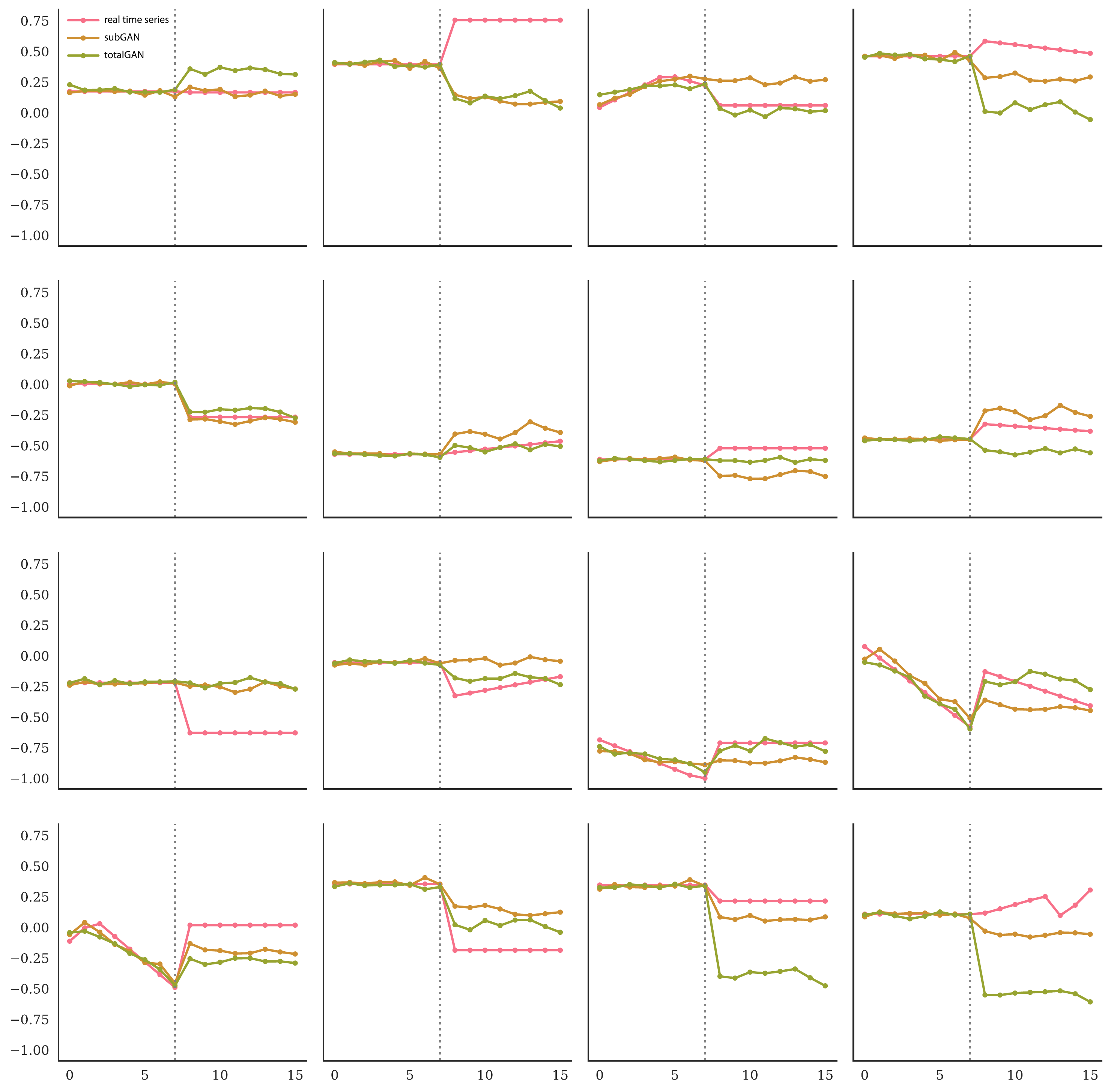}
	\caption*{Figure S4. Cluster 1.}
\end{figure}

\begin{figure}[h]
	\centering
	\includegraphics[scale=0.50]{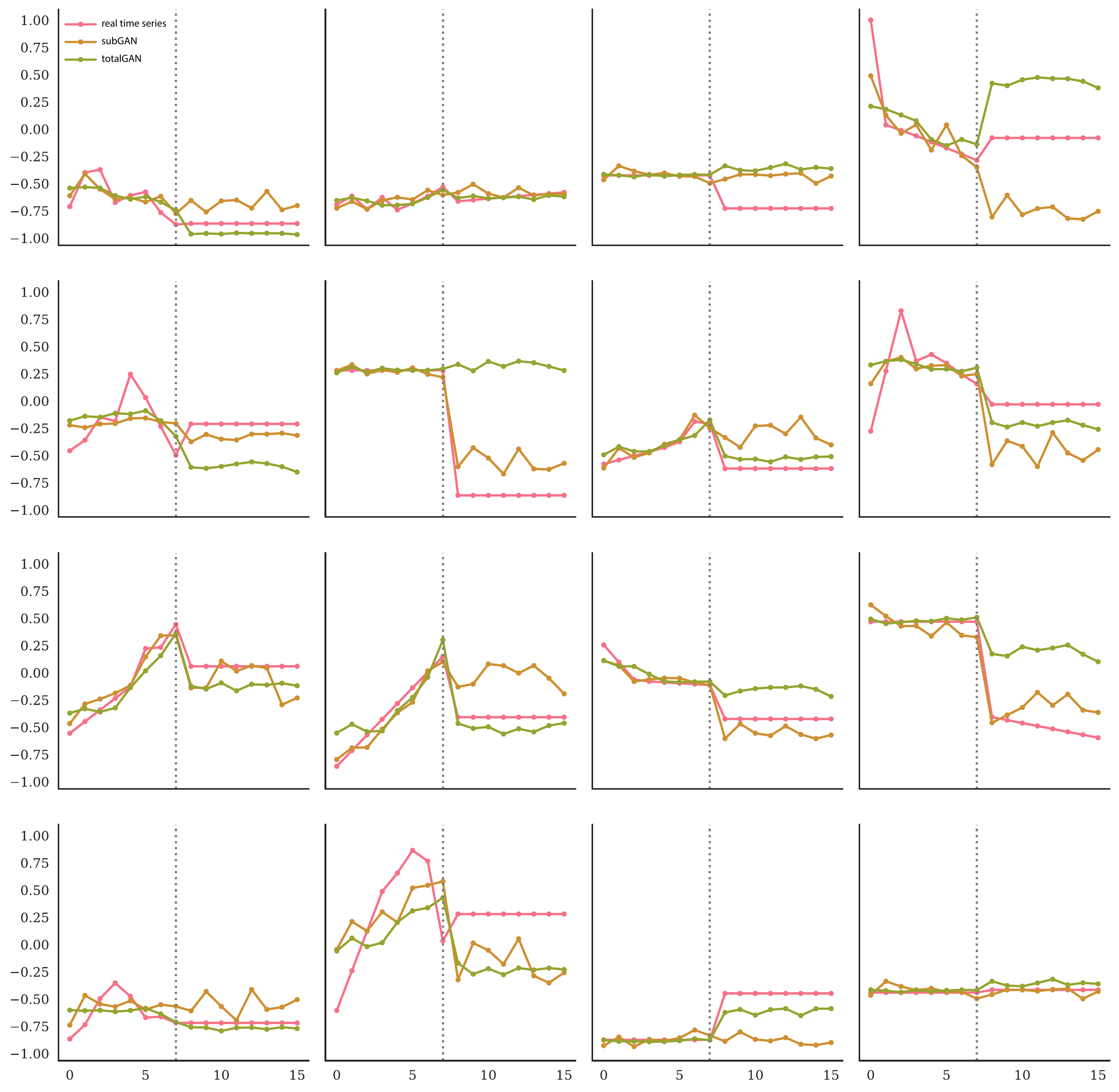}
	\caption*{Figure S5. Cluster 2.}
\end{figure}

\begin{figure}[h]
	\centering
	\includegraphics[scale=0.50]{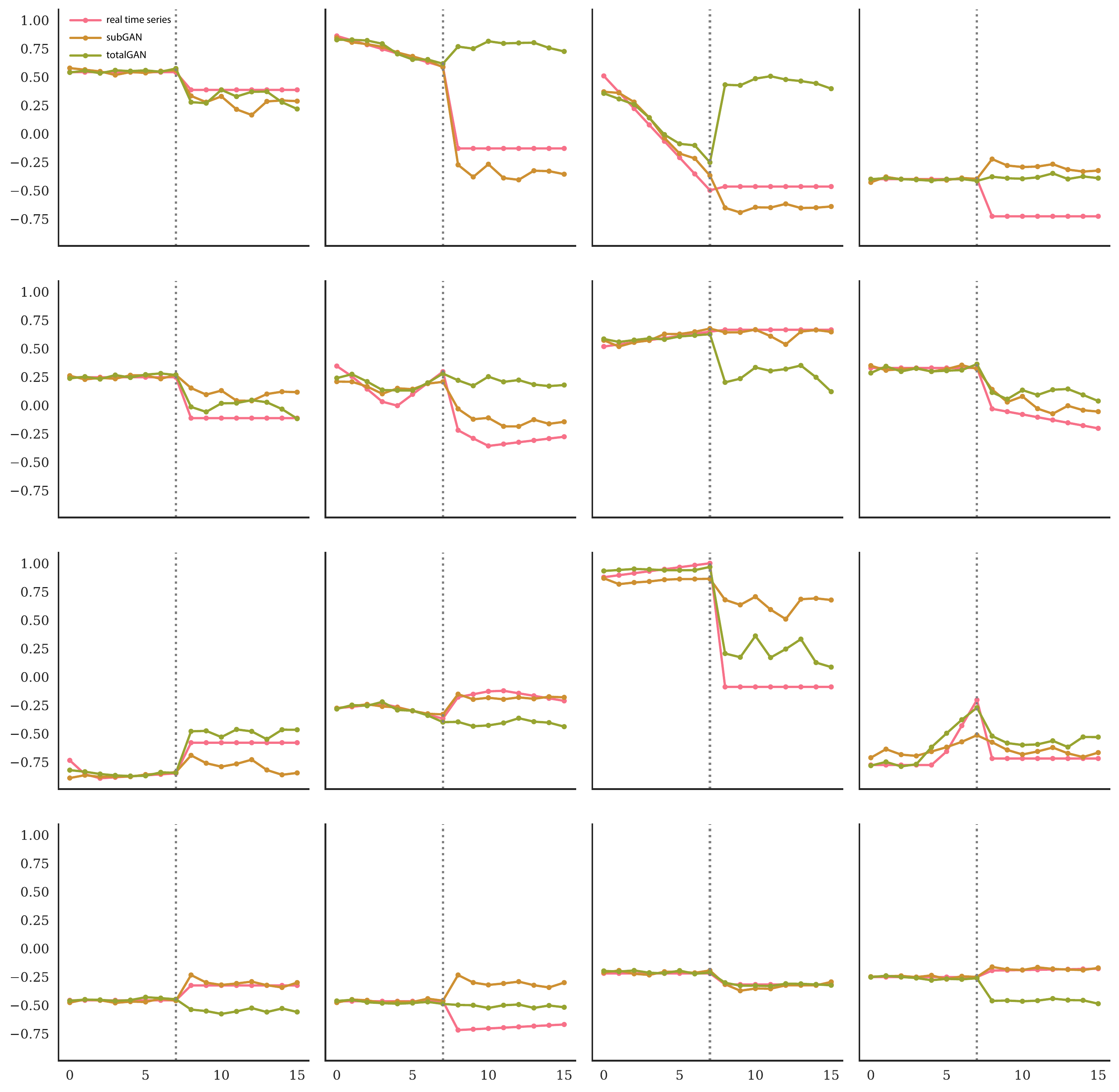}
	\caption*{Figure S6. Cluster 3.}
\end{figure}

\end{document}